%% file: main.tex
\documentclass[letterpaper,10pt,conference]{ieeeconf}
\IEEEoverridecommandlockouts
\usepackage{graphicx}
\usepackage{amsmath,amssymb}
\usepackage{booktabs}
\usepackage{cite}
\usepackage{url}
\usepackage{siunitx}
\makeatletter
\def\table{\def\@floatboxreset{\reset@font\footnotesize\@setminipage\vspace*{5pt}}\@float{table}}
\makeatother
\usepackage{balance}
\title{FDL: Foot--Ground Distance Layer for \\ Initial-Contact Load Modulation in a Lower-Limb Exoskeleton}

\author{Daeho Lee, Jeongsu Park, Hyunjin Choi, and Kyoungchul Kong%
\thanks{This research was supported by the National R\&D Program through the National Research Foundation of Korea (NRF) funded by the Ministry of Science and ICT (2020M3C1B8A01104835). \textit{(Corresponding authors: Hyunjin Choi; Kyoungchul Kong.)}}%
\thanks{Daeho Lee is with the 2nd R\&D Institute, Agency for Defense Development (ADD), Daejeon 34060, South Korea (e-mail: dhlee07042@add.re.kr).}%
\thanks{Jeongsu Park is with the Physical AI Business Unit, Angel Robotics, Daejeon 34327, South Korea (e-mail: jeongsu.park@angel-robotics.com).}%
\thanks{Hyunjin Choi is with the Dept. of Human Intelligence and Robot Engineering, Sangmyung University, Cheonan 31066, South Korea (e-mail: hyunjin@smu.ac.kr).}%
\thanks{Kyoungchul Kong is with Angel Robotics, Daejeon 34327, South Korea, and also with the Dept. of Mechanical Engineering, KAIST, Daejeon 34141, South Korea (e-mail: kckong@kaist.ac.kr).}
}

\begin{document}
\maketitle
\thispagestyle{empty}
\pagestyle{empty}

\begin{abstract}
In lower-limb exoskeletons, human movement variability can cause a wearer--robot timing mismatch that increases ground impact at initial contact. Existing controllers modulate assistance via estimated terrain, locomotion mode, or gait phase, rarely using foot--ground distance directly in the control law. This paper integrates direct foot--ground distance feedback into a spring-loaded inverted pendulum (SLIP)-based controller through a foot--ground distance layer (FDL). FDL uses a compact foot-mounted module with four time-of-flight sensors and applies a bounded shortening to the reference leg length. A single-participant pilot study was conducted on a tandem force-plate treadmill across successive FDL off, weak-gain, and strong-gain trials. Relative to the off trial, knee flexion at heel contact was 5.1\si{\degree} and 9.6\si{\degree} greater in the weak and strong trials. Peak anterior-plate force over the first \SI{50}{ms} was 9.0\% and 15.9\% lower, and impulse over the first \SI{100}{ms} was 24.7\% and 28.9\% lower. FDL used foot--ground distance to adjust the reference leg length before contact, and both active trials showed lower early vertical ground reaction force of the leading limb than the off trial.
\end{abstract}

\begin{keywords}
Prosthetics and Exoskeletons, Wearable Robotics, Physical Human-Robot Interaction, Range Sensing, Time-of-Flight Sensors.
\end{keywords}

\begin{figure*}[t]
\centering
\includegraphics[width=0.97\textwidth]{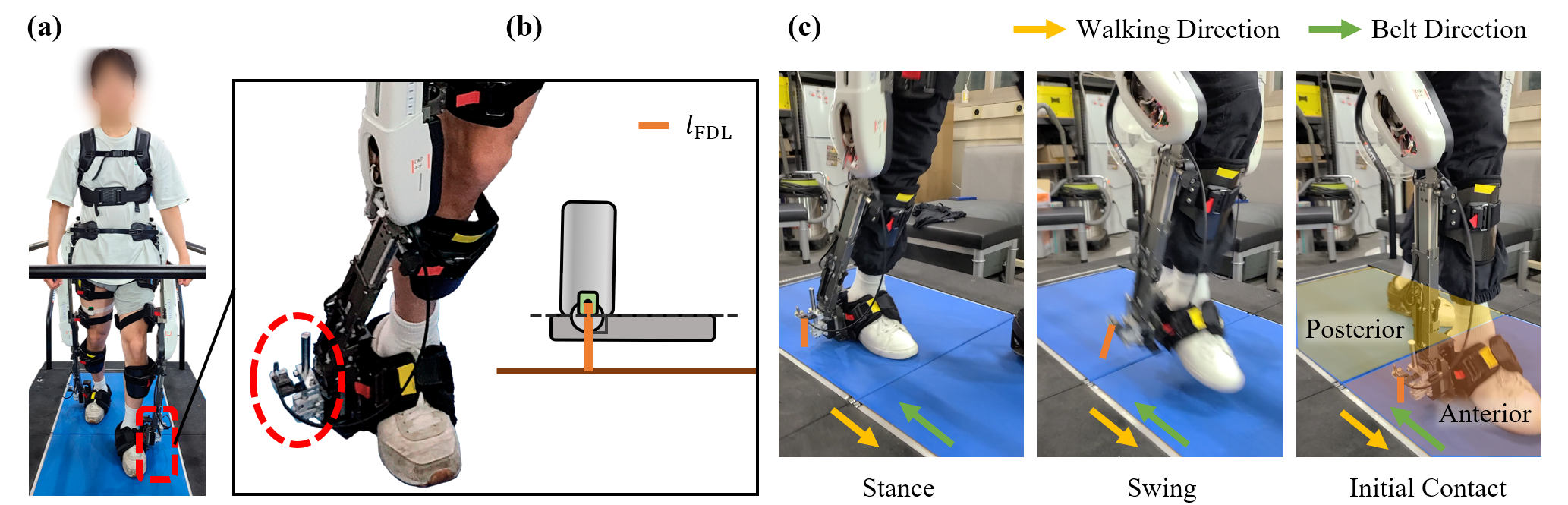}
\caption{FDL hardware and experimental geometry. (a) Participant wearing the exoskeleton on the tandem force-plate treadmill. (b) Close-up and ranging schematic of the foot-mounted four-ToF module measuring \(l_{\mathrm{FDL}}\) to the ground. (c) Force-plate assignment: the trailing foot stays on the posterior plate while the leading foot lands on the anterior plate beyond the boundary; consecutive contacts alternate between the feet.}
\label{fig:overview}
\end{figure*}

\section{Introduction}
A lower-limb exoskeleton is physically coupled to its wearer, and every step ends with the foot striking the ground. The vertical ground reaction force (vGRF) rises sharply at initial contact, transmitting impact forces to both the wearer and the robot. Since step-to-step variability in human movement cannot be fully predicted, the controller is unable to determine exactly when the foot will touch the ground, making impact difficult to mitigate before contact occurs.

Over repeated steps, large impact forces can cause wearer discomfort or injury while placing mechanical stress on the hardware. To reduce these impacts, prior work has proposed mechanical shock absorption~\cite{Park2021Impact} and joint-trajectory modulation~\cite{Zhang2022Trajectory}. The mechanical approach uses a spring--damper mechanism on the shank to take up the impact after the foot has landed. It adds mass and mechanical complexity to the leg, and its stiffness and damping are difficult to adjust online. The trajectory modulation introduces virtual spring--damper compliance through the knee trajectory of a position-controlled exoskeleton following initial contact. It requires no additional hardware, but it activates only after the strike has been detected and assumes that the joints track a prescribed trajectory. Neither approach prepares the leg before the foot touches the ground.

Recent wearable-robot perception and control methods do look ahead, though for different purposes. Vision-based systems classify upcoming terrain to modulate hip-exosuit assistance~\cite{Tricomi2023Environment,Tricomi2025Geometric}. Other perception frameworks map the surrounding scene~\cite{Wang2025Exosense}, recognize locomotion modes~\cite{Shin2025Multimodal}, or forecast the timing and center of pressure of upcoming foot contact~\cite{Murray2026Forecast}. Separately, user-adaptive controllers adjust assistance through real-time gait-phase and task estimation~\cite{Medrano2023Realtime}, human-in-the-loop optimization~\cite{Wang2025HILO}, or end-to-end learned policies~\cite{Barati2026EndToEnd}. Across these controllers, assistance is tailored to the terrain, task, or wearer via high-level estimates, parameter optimization, or learned mappings, while the measured distance between the swing foot and the ground rarely enters the control law directly.

Local distances to the terrain or ground can be measured directly with compact non-contact range sensors, which are already used in wearable robotics. A waist-mounted laser distance meter combined with an inertial sensor classified upcoming terrain for a powered prosthesis~\cite{Liu2016Environment}. Shoe-mounted infrared sensors detected gait events in a lower-limb exoskeleton~\cite{Li2021IRGait}, and foot-mounted time-of-flight (ToF) sensors estimated foot clearance~\cite{Jacob2021Clearance}. Foot-mounted ToF measurements of obstacle and ground clearance have also driven a virtual potential field that modifies the swing trajectory of an exoskeleton to prevent tripping~\cite{Zhang2020Tripping}, without addressing the force at initial contact.

This paper proposes the foot--ground distance layer (FDL), an add-on layer for an existing controller. FDL takes the foot--ground distance measured throughout the gait cycle by a compact foot-mounted module of four interleaved ToF sensors and outputs a bounded shortening of the reference leg length in a spring-loaded inverted pendulum (SLIP)-based controller. Shortening the reference leg shifts assistance toward greater knee flexion, a posture associated with lower peak vertical impact forces in running~\cite{Derrick2004KneeAngle}. Because FDL modifies a single reference variable, it requires no terrain classifier, contact forecaster, or trajectory planner while remaining compatible with higher-level adaptation methods. FDL is evaluated in a single-participant, fixed-order pilot study on a tandem force-plate treadmill (one trial each for FDL off, weak FDL gain, and strong FDL gain), analyzing both anterior-plate vGRF and exoskeleton controller data.

The contributions of this paper are as follows:
\begin{itemize}
\item a compact foot-mounted module with four ToF sensors that measures foot--ground distance throughout the gait cycle;
\item FDL, which feeds this distance directly into the SLIP-based control law of a powered lower-limb exoskeleton via a bounded shortening of the reference leg length; and
\item a single-participant treadmill pilot study across three successive trials (FDL off, weak, and strong) characterizing reduced peak and impulse of the leading-limb vGRF, right-knee flexion at heel contact, and controller saturation in swing and stance.
\end{itemize}

\section{FDL System and Control Integration}
\subsection{Foot-Mounted Distance Module}
\begin{figure}[t]
\centering
\includegraphics[width=\columnwidth]{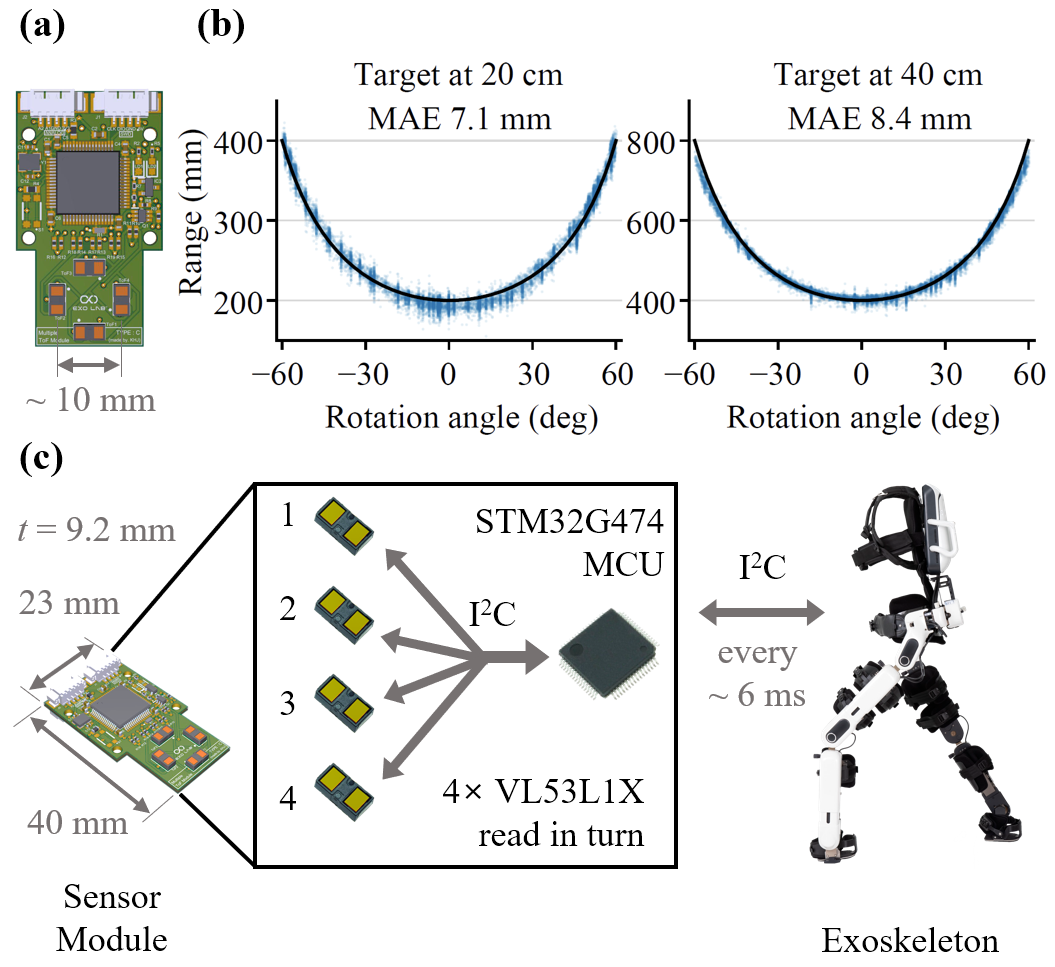}
\caption{Foot-mounted distance module. (a) Custom PCB with four VL53L1X sensors. (b) Bench ranging over \ang{\pm 60} at target distances \(d\): geometric reference \(d/\cos\theta\) (black), interleaved single-sensor readings (blue), and mean absolute error (MAE). (c) Interleaved readout and I\textsuperscript{2}C link to the exoskeleton's per-limb interface board (overall thickness \(t=\SI{9.2}{mm}\) across front and back connectors).}
\label{fig:sensor}
\end{figure}

Each distance module carries four VL53L1X direct ToF sensors placed within an approximately \SI{10}{mm} square on one board, with its own STM32G474 microcontroller (Fig.~\ref{fig:sensor}). The four sensors face the same direction and range nearly the same ground area; they are interleaved to raise the update rate, and \(l_{\mathrm{FDL}}\) is the most recent single-sensor reading.
The original module and preliminary walking data were reported in D. Lee's master's thesis at KAIST~\cite{Lee2022Thesis}; the bench-error computation, initial-contact force reanalysis, sensitivity grid, and reconstructions from the exoskeleton logs presented here are new.
In a bench test, the module was rotated from \ang{-60} to \ang{60} in front of a flat target at \SI{20}{cm} and \SI{40}{cm} (Fig.~\ref{fig:sensor}b). After correcting a constant zero-angle offset in each run, the mean absolute range error was 7.1~mm (SD 5.3~mm) at \SI{20}{cm} and 8.4~mm (SD 8.2~mm) at \SI{40}{cm}.

The microcontroller sets each sensor to short-distance mode with a \SI{10}{ms} timing budget, a configuration in which one sensor ranges at up to \SI{100}{Hz}~\cite{ST2019AN5263}. It then polls the four sensors in a fixed rotation, nominally reading and restarting the next sensor every \SI{6}{ms}. Each sensor is revisited every \SI{24}{ms} (\SI{41.7}{Hz}), giving an interleaved read schedule of \SI{167}{Hz} without averaging the array. A per-limb sensor interface board retrieves the latest reading and its range status over I\textsuperscript{2}C and forwards them to the main controller. In the recorded distances, the four-sensor pattern recurs approximately every 4.8 samples at the nominal \SI{5}{ms} logging interval (an estimated rotation tick of 5.97--6.03~ms), consistent with the \SI{6}{ms} rotation.

Each module is mounted on the lateral foot frame near the ankle, pointing downward behind the heel so that it rotates with the foot; \(l_{\mathrm{FDL}}\) is therefore a distance to the ground that depends on foot orientation rather than vertical sensor height or sole clearance.

\begin{figure*}[t]
\centering
\includegraphics[width=0.97\textwidth]{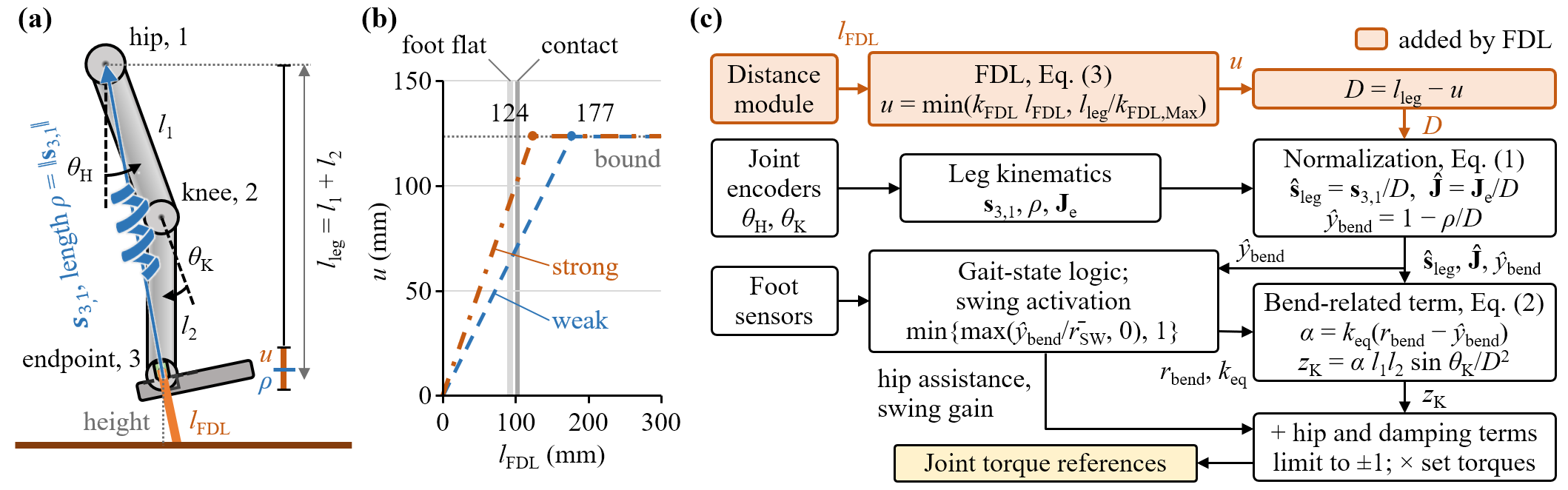}
\caption{FDL integration into the SLIP-based controller. (a) Sagittal leg model: links \(l_1\) and \(l_2\), hip and knee angles, and the leg vector \(\mathbf{s}_{3,1}\) from the foot-side endpoint (3) to the hip (1), of length \(\rho\). The module rotates with the foot, so \(l_{\mathrm{FDL}}\) is a slant distance, not vertical sensor height. (b) Shortening \(u\) of~\eqref{eq:bound} for the two settings (\(l_{\mathrm{leg}}=\SI{990}{mm}\), \(k_{\mathrm{FDL,Max}}=8\)), which reach the bound at 124~mm (strong) and 177~mm (weak); shaded, the foot-flat and at-contact distances recorded in the trials. (c) Data flow in the controller; orange-shaded blocks are added by FDL.}

\label{fig:controller}
\end{figure*}

\subsection{SLIP-Based Assistance and FDL Modification}
The exoskeleton, Angel Legs M20 (AL~M20; Angel Robotics, Seoul, South Korea), actuates the hip and knee of each leg (the ankle is passive) and computes their torque references from a gait-state estimator and a SLIP-based assistance model~\cite{Blickhan1989SpringMass}. Let \(\mathbf{s}_{3,1}=-\mathbf{p}_{\mathrm{e}}\) be the vector from the foot-side endpoint (point~3 in Fig.~\ref{fig:controller}a), slightly below the ankle, to the hip (point~1), \(\rho=\|\mathbf{s}_{3,1}\|\), \(\mathbf{J}_{\mathrm{e}}=\partial\mathbf{p}_{\mathrm{e}}/\partial\mathbf{q}\) the Jacobian of the hip-to-endpoint position with respect to the hip and knee angles \(\mathbf{q}=(\theta_{\mathrm{H}},\theta_{\mathrm{K}})\), and \(l_{\mathrm{leg}}=l_1+l_2\). The controller normalizes the leg by a length \(D\):
\begin{equation}
 \hat{\mathbf{s}}_{\mathrm{leg}}=\frac{\mathbf{s}_{3,1}}{D},\hspace{0.75em}
 \hat{\mathbf{J}}=\frac{\mathbf{J}_{\mathrm{e}}}{D},\hspace{0.75em}
 \hat{y}_{\mathrm{bend}}=1-\|\hat{\mathbf{s}}_{\mathrm{leg}}\|=1-\frac{\rho}{D}.
\end{equation}
Without FDL, \(D=l_{\mathrm{leg}}\), and the bend \(\hat{y}_{\mathrm{bend}}\) is the fraction by which the leg is shorter than its full length. In its unconstrained form, the bend-related term scales the normalized leg vector by \(\alpha=k_{\mathrm{eq}}(r_{\mathrm{bend}}-\hat{y}_{\mathrm{bend}})\), where \(r_{\mathrm{bend}}\) is a bend reference that the gait-state logic updates (in stance, from \(\hat{y}_{\mathrm{bend}}\)) and \(k_{\mathrm{eq}}\) a state-dependent gain, and maps it to a joint command \(\mathbf{z}=\mathbf{S}\hat{\mathbf{J}}^{\mathsf T}\alpha\hat{\mathbf{s}}_{\mathrm{leg}}\), with \(\mathbf{S}\) the fixed sign convention of the command (positive, flexion). Since \(\alpha=k_{\mathrm{eq}}[\rho-(1-r_{\mathrm{bend}})D]/D\), the term acts as a nonlinear virtual spring along the hip--endpoint line that drives \(\rho\) toward \((1-r_{\mathrm{bend}})D\). Because this vector lies on the line through the hip (\(\mathbf{J}_{\mathrm{e}}^{\mathsf T}\mathbf{s}_{3,1}=[0,\;l_1l_2\sin\theta_{\mathrm{K}}]^{\mathsf T}\)), \(\mathbf{z}\) has no hip component; its knee component is
\begin{equation}
 z_{\mathrm{K}}=\frac{\alpha\,l_1l_2\sin\theta_{\mathrm{K}}}{D^{2}},
 \label{eq:torque}
\end{equation}
where \(\theta_{\mathrm{K}}\) is the knee flexion angle, so \(r_{\mathrm{bend}}>\hat{y}_{\mathrm{bend}}\) commands knee flexion. The controller adds hip and damping terms to this normalized command, limits the sum to \(\pm1\), and scales it by set flexion and extension torques. FDL was designed to bias the knee toward flexion during swing in preparation for upcoming contact. It shortens the normalization length, which serves as the reference leg length, using the module distance \(l_{\mathrm{FDL}}\ge0\):
\begin{equation}
 D=l_{\mathrm{leg}}-u,\hspace{0.75em}
 u=\min\!\bigl(k_{\mathrm{FDL}}\,l_{\mathrm{FDL}},\,l_{\mathrm{leg}}/k_{\mathrm{FDL,Max}}\bigr).
 \label{eq:bound}
\end{equation}
 The bound \(u_{\max}=l_{\mathrm{leg}}/k_{\mathrm{FDL,Max}}\) applies to the shortening, not to the distance: \(u\) saturates for \(l_{\mathrm{FDL}}\ge u_{\max}/k_{\mathrm{FDL}}\), so a higher gain reaches the same limit at a shorter distance (Fig.~\ref{fig:controller}b). With the settings used here, \(u\) therefore varies with distance only below 177~mm (weak) and 124~mm (strong), that is, only when the foot is near the ground; below this saturation threshold, the shortening tapers. Holding posture, reference, and a positive \(k_{\mathrm{eq}}\) fixed, the shorter \(D\) lowers the target length \((1-r_{\mathrm{bend}})D\), lowers \(\hat{y}_{\mathrm{bend}}\) (below zero once \(\rho>D\)), and enlarges the factor \(1/D^2\) in~\eqref{eq:torque}; together these shift \(z_{\mathrm{K}}\) toward flexion whenever \(3\rho>2(1-r_{\mathrm{bend}})\,l_{\mathrm{leg}}\).

FDL also acts beyond the knee term. The same \(\hat{y}_{\mathrm{bend}}\) sets the swing activation, \(\min\{\max(\hat{y}_{\mathrm{bend}}/\bar{r}_{\mathrm{SW}},0),1\}\) with \(\bar{r}_{\mathrm{SW}}\) the swing-bend setting, which contributes to the swing-state logic and scales the hip assistance and the swing gain. FDL can therefore change the timing and the hip command as well.

The controller uses \(l_{\mathrm{FDL}}\) directly without rejecting readings based on the ToF sensor range status. Because \(0 \le u \le u_{\max}\), any ranging error keeps \(D\) between \(l_{\mathrm{leg}}-u_{\max}\) and \(l_{\mathrm{leg}}\): a spurious short reading moves \(D\) toward the FDL-off value \(l_{\mathrm{leg}}\), and a spurious long reading saturates it at \(l_{\mathrm{leg}}-u_{\max}\). When the foot reaches the ground, the module's mounting height above the sole keeps \(l_{\mathrm{FDL}}\) in the nonzero foot-flat range of 88--97~mm (Sec.~\ref{sec:exoresults}), so \(u\) remains nonzero during stance.

\section{Human-Participant Pilot Evaluation}
\subsection{Protocol and Controller Settings}
One healthy male participant in his twenties (height \SI{182}{cm}, mass \SI{78}{kg}) walked at \SI{1.5}{km/h} (\SI{0.42}{m/s}) on an instrumented treadmill with two tandem force plates (AMTI, Watertown, MA, USA) while wearing AL~M20. A safety handrail surrounded the treadmill; the participant did not hold it during the recorded trials, and no harness or crutches were used. The distance modules were mounted bilaterally, and FDL was applied to both legs; all other assistance settings of the exoskeleton were the same in every condition. The experiment comprised three trials in a fixed order: FDL off, weak FDL gain, and strong FDL gain (Table~\ref{tab:protocol}). Because only one trial was collected per condition, the two active trials are analyzed as discrete settings rather than as points on a gain--response curve.
The protocol was approved by the KAIST Institutional Review Board (IRB No.~\mbox{KH2020-162}). The participant received a detailed explanation of the experiment and gave his consent.

\input{tables/table_protocol.tex}

\subsection{Force-Plate Assignment and Event Definition}
To separate leading- and trailing-limb forces across the tandem plates, the participant was instructed to keep the trailing foot on the posterior plate and land the leading foot on the anterior plate (Fig.~\ref{fig:overview}c). Before recording, the participant practiced this foot placement during a familiarization period until reaching a steady gait. Left and right feet alternated on the anterior plate, as confirmed by mediolateral center-of-pressure shifts across the plate midline in at least 97\% of consecutive contacts per trial. Over the first \SI{100}{ms} of contact, the posterior plate carried mainly the trailing limb in late stance, recording a mean vertical force of 749--801~N across conditions versus 181--255~N on the anterior plate. Summing both plates would add this trailing-limb vertical force to the leading-limb signal, so the primary outcomes used the anterior plate alone. Hereafter, vGRF refers to the anterior plate (leading limb) unless stated otherwise.

Candidate contacts were detected as unloaded-to-loaded transitions of the vGRF after low-pass Butterworth filtering (zero-phase, second-order, \SI{25}{Hz}). Force onset was defined for each candidate as the first upward crossing of \SI{20}{N} in the raw \SI{1}{kHz} signal, a vertical-force threshold also reported for force-plate gait-event detection in~\cite{Vancanneyt2025GaitEvents}. At the start of each recording, the first candidate (boundary artifact) and next three contacts (start-up transients) were discarded, along with the final candidate; the retained contact counts in Table~\ref{tab:protocol} reflect differences in trial duration and step interval. Two checks verified the plate assignment: between contacts, the vGRF returned to or below \SI{0}{N} while the posterior-plate force remained at or above \SI{50}{N} (138 of 139 intervals), and the mean vGRF over the \SI{20}{ms} preceding each candidate was within 2.9~N of zero.

\begin{figure*}[t]
\centering
\includegraphics{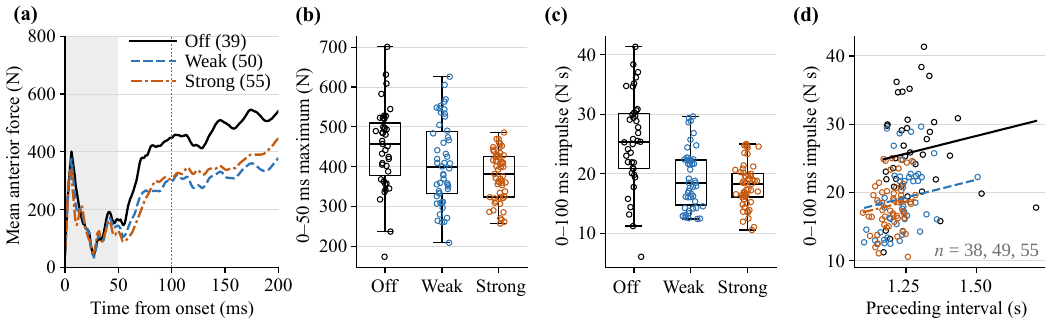}
\caption{Ground impact at initial contact (anterior plate). (a) Ensemble-mean vGRF aligned to the force onset (retained contact counts in parentheses; shaded band, \num{0}--\SI{50}{ms} window; dotted line, \SI{100}{ms} impulse boundary). (b) Peak vGRF within \num{0}--\SI{50}{ms} and (c) vGRF impulse over \num{0}--\SI{100}{ms} for all retained contacts (boxes, median and interquartile range [IQR]; whiskers, up to \(1.5\times\text{IQR}\); markers, individual contacts). (d) vGRF impulse versus the preceding step interval (intervals \(\le\SI{1.8}{s}\); counts for Off, Weak, and Strong shown in panel), with common-slope linear fits.}
\label{fig:force}
\end{figure*}

Ground impact was quantified by two metrics computed from the unfiltered signal: the peak vGRF within \num{0}--\SI{50}{ms} of onset and the vGRF impulse (the time integral of the positive vGRF) over \num{0}--\SI{100}{ms}. The impulse was also split at \SI{50}{ms} to determine when differences occurred. Later stance was not analyzed because the belt carries the foot rearward across the plate boundary. Because all contacts were collected from a single participant, repeated steps are not treated as independent replications of a condition.

\subsection{Sensitivity Analyses}
Robustness to data-processing choices was verified by recomputing the outcomes across \num{54} parameter combinations: three onset thresholds (\SIlist{10;20;50}{N}), three detection-filter cutoffs (\SIlist{25;50;100}{Hz}), three initial-step exclusions (0, 3, or 5 steps), and inclusion or exclusion of steps following inter-contact gaps above \SI{1.8}{s}. Outcomes were also computed separately for each limb. To assess sensitivity to high-frequency transients, the peak vGRF was recomputed for the primary contacts and onsets after zero-phase, second-order Butterworth low-pass filtering at \SI{50}{Hz} and \SI{25}{Hz}.

Two additional checks examined step timing and within-trial drift across the fixed trial order in Table~\ref{tab:protocol}. On the gap-filtered subset (preceding interval \(\le\SI{1.8}{s}\)), the \num{0}--\SI{100}{ms} impulse was regressed linearly on the preceding step interval with a common slope and one intercept per condition; this descriptive model tests sensitivity to step timing, although it cannot separate controller settings from trial-order effects. Each trial's retained contacts were also split into first and second halves to check for within-trial drift in the \num{0}--\SI{100}{ms} impulse.

\subsection{Exoskeleton Data}
The exoskeleton recorded joint angles, joint commands, foot-sensor signals, and each module's distance and range status at a nominal \SI{5}{ms} interval. The exoskeleton log corresponding to each force-plate trial was identified by comparing right heel-contact sequences across candidate logs. The selected pairs had RMS timing residuals of 24--37~ms over 22--30 strides, compared with at least 71~ms for alternative pairings.

Force-plate and exoskeleton outcomes were analyzed separately. The vGRF outcomes were computed from the \SI{1}{kHz} force-plate recordings. The exoskeleton measurements were expressed as a percentage of each leg's gait cycle relative to its segmented initial contact. Offline gait cycles were segmented using the exoskeleton's heel sensor on the right leg and the module distance on the left leg because the left heel-sensor signal was unreliable. On the right leg, initial contact was defined as an upward threshold crossing of the heel-sensor signal; results that depend on this event were repeated at two higher thresholds. The shortening \(u\) was reconstructed offline from the recorded distance using~\eqref{eq:bound} with configured link lengths \(l_1=\SI{470}{mm}\), \(l_2=\SI{520}{mm}\), and \(k_{\mathrm{FDL,Max}}=8\); stance and swing samples were classified from the right heel-sensor signal.

\section{Results}
\subsection{Ground Impact}
Mean anterior-plate peak force within the first \SI{50}{ms} was 448.8~N in the FDL off trial, 408.5~N in the weak trial, and 377.3~N in the strong trial (Table~\ref{tab:metrics}, Fig.~\ref{fig:force}). Individual peaks typically occurred 5--6~ms after force onset. Over the first \SI{100}{ms}, anterior impulse was 24.7\% and 28.9\% lower than in the off trial, respectively. Most of this impulse difference accumulated after \SI{50}{ms}: between \num{50} and \SI{100}{ms}, mean impulse was 16.97, 11.30, and 10.73~N\,s, accounting for 90\% (weak) and 85\% (strong) of the \num{0}--\SI{100}{ms} differences.

\input{tables/table_metrics.tex}

To check whether reduced leading-limb force was offset by greater trailing-limb vGRF, the posterior plate was examined over the same \num{0}--\SI{100}{ms} window. Because the cycle-mean summed vGRF was approximately 53~N lower in the weak trial than in FDL off (while strong FDL closely matched FDL off), two-plate vGRF was evaluated both before and after subtracting each trial's cycle-mean summed force (Table~\ref{tab:metrics}). With strong FDL, posterior-plate vGRF exceeded FDL off under both conventions; with weak FDL, trial-mean subtraction reversed the difference from negative to positive. Nevertheless, the combined two-plate impulse stayed lower in both active trials regardless of subtraction.

\begin{figure}[t]
\centering
\includegraphics{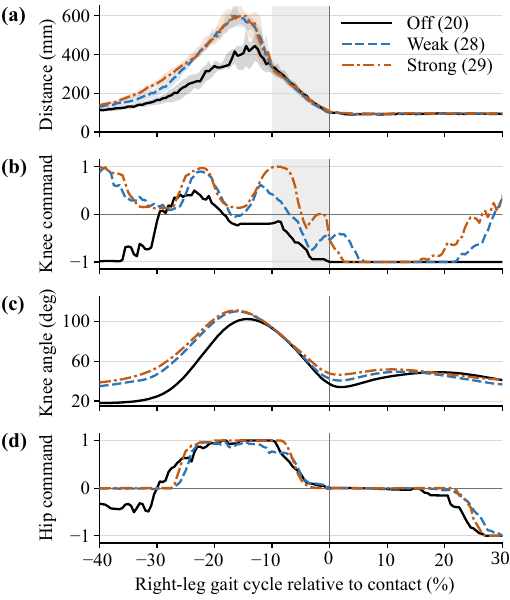}
\caption{Right-leg exoskeleton data around initial contact (\SI{0}{\%}), in percent of the gait cycle (stride counts in parentheses). (a) Module distance \(l_{\mathrm{FDL}}\), median and IQR of readings with zero range status. (b) Mean knee command normalized by its saturation limits (\(+1\), flexion; \(-1\), extension). (c) Mean knee flexion angle \(\theta_{\mathrm{K}}\) from the joint encoder. (d) Mean hip command, normalized as in (b). Vertical shading in (a, b), last 10\% of the cycle before contact.}
\label{fig:robot}
\end{figure}

\subsection{Assistance Commands and Joint Angles}
\label{sec:exoresults}
The distance modules read 88--97~mm with the foot flat and 100--106~mm at initial contact (Fig.~\ref{fig:robot}a). Readings with a nonzero range status occurred mainly above \SI{400}{mm} in mid-swing.

Before contact (\(-10\) to \SI{0}{\%} of the gait cycle), FDL shifted the normalized right-leg knee torque command (knee command) toward flexion, progressing from extension with FDL off to flexion with strong FDL (Table~\ref{tab:robot}, Fig.~\ref{fig:robot}b). At initial contact (heel contact), right-knee flexion was 5.1\si{\degree} greater in the weak trial and 9.6\si{\degree} greater in the strong trial than with FDL off (Fig.~\ref{fig:robot}c). Both patterns held across all three contact thresholds, with contact flexion remaining 5.1--6.5\si{\degree} and 9.6--11.3\si{\degree} greater. By heel contact, however, the knee command had returned to extension in all three trials and remained strongly negative in early stance (\num{0} to \SI{10}{\%}; Table~\ref{tab:robot}).

Stance-phase commands and kinematics also differed in the active trials. Rising in mid-to-late stance, the right-leg knee command saturated at its flexion limit in 37\% (weak) and 39\% (strong) of stance samples, versus 0\% with FDL off (Fig.~\ref{fig:robot}b, Table~\ref{tab:robot}). At peak extension, the right knee remained 15.4\si{\degree} and 20.0\si{\degree} more flexed than with FDL off (Fig.~\ref{fig:robot}c, Table~\ref{tab:robot}), with the left leg showing the same ordering. The hip command changed less (Fig.~\ref{fig:robot}d). Its fraction of stance samples at the extension limit stayed similar across trials (26--33\%), although late-stance extension assistance (\(-40\) to \SI{-30}{\%}; Table~\ref{tab:robot}) was absent under FDL, and both swing-flexion and stance-extension onsets occurred slightly later.

Reconstructing \(u\) showed when the distance modulated the reference leg length (Table~\ref{tab:robot}). Throughout most of swing, \(u\) remained saturated at \(u_{\max}\)---in 61\% (weak) and 86\% (strong) of right-leg heel-unloaded samples---and exited saturation at medians of 4.0\% and 0.8\% of the gait cycle before contact, respectively. Between \(-10\) and \SI{-4}{\%}, \(u\) stayed at \(u_{\max}\) in 95\% (weak) and 100\% (strong) of samples while the window-averaged right-knee angles differed by 1.0\si{\degree} (with the mean curves diverging toward the end of the window; Fig.~\ref{fig:robot}c), yet the mean knee command still differed (0.04 vs.\ 0.69). During stance (\num{5} to \SI{30}{\%}), \(u\) stayed near 67 and 96~mm, respectively, reflecting the nonzero foot-flat module offset.

\input{tables/table_robot.tex}

\subsection{Robustness and Within-Trial Checks}
Anterior-plate peak vGRF and impulse remained below FDL off across all \num{54} processing variants, with the weak-trial peak showing the widest spread (\(-13.4\) to \(-4.9\)\%; Table~\ref{tab:sens}). Both feet showed lower vGRF when analyzed separately, though peak differences varied between limbs. Low-pass filtering at \SI{50}{Hz} and \SI{25}{Hz} enlarged the relative peak differences. Accounting for step-interval variation with the common-slope model (\(\hat{\beta}=+10.4\)~N\,s/s; Fig.~\ref{fig:force}d) shifted the impulse differences only slightly. Within every trial, however, the mean \num{0}--\SI{100}{ms} impulse rose from the first half to the second.

\section{Discussion}
Limited update rates have been a practical barrier to using ToF measurements directly in the control law: Jacob et al.~\cite{Jacob2021Clearance} recorded foot-clearance data at \SI{50}{Hz}, and a single VL53L1X caps at \SI{100}{Hz}~\cite{ST2019AN5263}. Interleaving four sensors raised the update rate to approximately \SI{167}{Hz}. Although this rate is still slower than typical \SI{1}{kHz} torque loops and unable to react to the 5--6~ms peak, FDL shapes a mid-level reference during terminal swing rather than forming a closed impact-mitigation loop after contact. Its benefit over slower polling and total latency at higher walking speeds remain to be studied. Because FDL acts prior to contact, it can serve as a complementary approach when integrated with post-contact mechanical or trajectory-based methods~\cite{Park2021Impact, Zhang2022Trajectory}.

The anterior vGRF in Fig.~\ref{fig:force}a shows a two-stage profile: an abrupt 5--6~ms peak, a brief dip, and a gradual rise after \SI{50}{ms}, separating an early collision transient from slow weight acceptance. One possible explanation is a brief rebound following collision between the exoskeleton sole and the treadmill surface. However, the exact mechanism of this profile cannot be verified because vertical plate force alone cannot isolate limb, frame, and deck-resonance contributions~\cite{Sloot2015Treadmill}, nor confirm reduced skeletal shock~\cite{Derrick2004KneeAngle}. Whether unsaturated terminal-swing tuning alters this vGRF profile remains to be tested.

Future work should address two main limitations. First, shortening \(u\) remained saturated through most of swing and nonzero in stance, so these trials cannot isolate live distance feedback from sustained shortening or carried-over controller state. Separating these effects requires experimentally comparing an offset-adjusted, unsaturated FDL against a predefined, phase-indexed shortening profile. Second, the single-participant, fixed-order level-walking protocol leaves the effects from trial order unresolved. Follow-up trials should be counterbalanced across multiple participants and higher speeds, incorporating unloaded treadmill zeroing~\cite{Sloot2015Treadmill} to control baseline offsets, and reliable bilateral foot sensing. In addition, wearer-response measures including EMG~\cite{Wang2025HILO} and metabolic cost, together with motion capture, would be required.

\input{tables/table_sensitivity.tex}

\section{Conclusion}
This paper presented FDL, which feeds foot--ground distance from an interleaved four-ToF module directly into a SLIP-based exoskeleton controller to shorten its reference leg length. In a single-participant, fixed-order pilot study, the active trials showed greater right-knee flexion at heel contact and lower early anterior-plate vGRF, with most of the \num{0}--\SI{100}{ms} impulse difference arising after \SI{50}{ms}. Exoskeleton log analysis showed that shortening remained saturated through most of swing and nonzero in stance, identifying the saturation threshold and foot-flat offset as key parameters for terminal-swing modulation.

\section*{Acknowledgment}
K.~Kong is a co-founder and the chief executive officer of Angel Robotics, the manufacturer of the exoskeleton used in this study.

\balance
\bibliographystyle{IEEEtran}
\bibliography{refs}
\end{document}

%% file: tables/table_protocol.tex
\begin{table}[t]
\caption{Controller settings, trial duration, mean step interval, and number of retained initial contacts for each trial.}
\label{tab:protocol}
\centering
\footnotesize
\setlength{\tabcolsep}{3.5pt}
\begin{tabular}{lccccc}
\toprule
Condition & \(k_{\mathrm{FDL}}\) & \(k_{\mathrm{FDL,Max}}\) & Duration (s) & Step (s) & Contacts \\
\midrule
FDL off & 0.0 & --- & 69.1 & 1.282 & 39 \\
Weak FDL & 0.7 & 8 & 78.3 & 1.237 & 50 \\
Strong FDL & 1.0 & 8 & 80.1 & 1.199 & 55 \\
\bottomrule
\end{tabular}
\end{table}

%% file: tables/table_metrics.tex
\begin{table}[t]
\caption{vGRF peak and impulse at initial contact with FDL off and with two FDL settings (mean\(\pm\)SD over retained contacts).}
\label{tab:metrics}
\centering
\footnotesize
\setlength{\tabcolsep}{2.2pt}
\begin{tabular}{lccc}
\toprule
Measure & FDL off & Weak & Strong \\
\midrule
\multicolumn{4}{@{}l}{\textit{Anterior plate (leading limb)}} \\
Peak vGRF, \(0\)--\SI{50}{ms} (N) & 448.8\(\pm\)103.6 & 408.5\(\pm\)104.0 & 377.3\(\pm\)61.2 \\
\quad change vs.\ off (\%) & --- & \(-9.0\) & \(-15.9\) \\
\addlinespace[3pt]
vGRF impulse, \(0\)--\SI{100}{ms} (N\,s) & 25.50\(\pm\)7.77 & 19.19\(\pm\)4.95 & 18.12\(\pm\)3.63 \\
\quad change vs.\ off (\%) & --- & \(-24.7\) & \(-28.9\) \\
\quad \(0\)--\SI{50}{ms} (N\,s) & 8.53\(\pm\)1.66 & 7.89\(\pm\)1.13 & 7.39\(\pm\)0.80 \\
\quad \(50\)--\SI{100}{ms} (N\,s) & 16.97\(\pm\)6.46 & 11.30\(\pm\)4.23 & 10.73\(\pm\)3.00 \\
\midrule
\multicolumn{4}{@{}l}{\textit{Both plates combined (\(0\)--\SI{100}{ms} impulse)}} \\
Summed impulse (N\,s) & 103.07\(\pm\)6.91 & 94.10\(\pm\)5.27 & 98.17\(\pm\)3.93 \\
Mean force removed (N\,s) & 14.14\(\pm\)6.91 & 10.43\(\pm\)5.27 & 9.65\(\pm\)3.93 \\
\quad difference vs.\ off (N\,s) & --- & \(-\)3.7 & \(-\)4.5 \\
\bottomrule
\end{tabular}
\end{table}

%% file: tables/table_robot.tex
\begin{table}[t]
\caption{Right-leg joint commands, knee angle, and shortening \(u\) (windows in \% gait cycle relative to initial contact; commands normalized to \(\pm1\), flexion positive; med.: median).}
\label{tab:robot}
\centering
\footnotesize
\setlength{\tabcolsep}{0pt}
\begin{tabular*}{\columnwidth}{@{\extracolsep{\fill}}lccc@{}}
\toprule
Quantity & FDL off & Weak & Strong \\
\midrule
\multicolumn{4}{@{}l}{\textit{Joint commands}} \\
\enspace Knee, \(-10\) to \SI{0}{\%} & \(-0.61\) & \(-0.21\) & 0.38 \\
\enspace Knee, \(-10\) to \SI{-4}{\%} & \(-0.40\) & 0.04 & 0.69 \\
\enspace Knee, 0 to \SI{10}{\%} & \(-1.00\) & \(-0.76\) & \(-0.93\) \\
\enspace Knee, stance at flexion limit (\% samples) & 0 & 37 & 39 \\
\addlinespace[3pt]
\enspace Hip, \(-40\) to \SI{-30}{\%} & \(-0.38\) & \(-0.01\) & 0.00 \\
\midrule
\multicolumn{4}{@{}l}{\textit{Knee angle (deg)}} \\
\enspace Flexion at contact vs.\ off & --- & 5.1 & 9.6 \\
\enspace Minimum per cycle, med. & 17.0 & 32.4 & 37.0 \\
\midrule
\multicolumn{4}{@{}l}{\textit{Shortening \(u\)}} \\
\enspace At \(u_{\max}\), heel unloaded (\% samples) & --- & 61 & 86 \\
\enspace At \(u_{\max}\), \(-10\) to \SI{-4}{\%} (\% samples) & --- & 95 & 100 \\
\enspace Exits \(u_{\max}\) before contact, med.\ (\% cycle) & --- & 4.0 & 0.8 \\
\enspace Over 5 to \SI{30}{\%}, med.\ (mm) & --- & 67 & 96 \\
\bottomrule
\end{tabular*}
\end{table}

%% file: tables/table_sensitivity.tex
\begin{table}[t]
\caption{Robustness of the analysis to processing parameters, foot, low-pass filtering, step interval, and time within the trial.}
\label{tab:sens}
\centering
\footnotesize
\setlength{\tabcolsep}{2.5pt}
\begin{tabular}{@{}lccc@{}}
\toprule
Check & FDL off & Weak & Strong \\
\midrule
\multicolumn{4}{@{}l}{\textit{Change vs.\ FDL off (\%)}} \\
\quad 54 combinations, impulse & --- & \(-27.6\) to \(-22.3\) & \(-30.3\) to \(-25.7\) \\
\quad 54 combinations, peak & --- & \(-13.4\) to \(-4.9\) & \(-17.6\) to \(-12.0\) \\
\addlinespace[3pt]
\quad Each foot, impulse & --- & \(-30\), \(-20\) & \(-30\), \(-28\) \\
\quad Each foot, peak & --- & \(-4\), \(-14\) & \(-22\), \(-9\) \\
\addlinespace[3pt]
\quad \SI{50}{Hz} low-pass, peak & --- & \(-\)14.2 & \(-\)24.9 \\
\quad \SI{25}{Hz} low-pass, peak & --- & \(-\)12.1 & \(-\)21.1 \\
\midrule
\multicolumn{4}{@{}l}{\textit{Impulse difference vs.\ FDL off (\(\le\SI{1.8}{s}\) intervals; N\,s)}} \\
\quad Unadjusted & --- & \(-6.88\) & \(-7.89\) \\
\quad Step-interval adjusted & --- & \(-6.42\) & \(-7.03\) \\
\midrule
\multicolumn{4}{@{}l}{\textit{Within-trial \(0\)--\SI{100}{ms} mean impulse (N\,s)}} \\
\quad First half & 23.90 & 18.33 & 16.91 \\
\quad Second half & 27.01 & 20.05 & 19.29 \\
\bottomrule
\end{tabular}
\end{table}